%% file: root.tex
\documentclass[letterpaper, 10 pt, conference]{ieeeconf}  %

\IEEEoverridecommandlockouts                              %

\usepackage{graphicx}
\usepackage{amsmath}
\usepackage{amssymb}
\usepackage{booktabs}
\usepackage{algorithm}
\usepackage{authblk}
\usepackage{algpseudocode}
\usepackage{listings}
\usepackage{subcaption}
\usepackage{wrapfig}
\usepackage{xspace, multirow, array}
\usepackage{hyperref}

\newcommand{\methods}{DiT-Block Policies\xspace}
\newcommand{\method}{DiT-Block Policy\xspace}
\newcommand{\datasetname}{BiPlay\xspace}

\newcommand{\ie}{i.e., }
\newcommand{\eg}{e.g., }

\newcommand{\website}{\url{https://dit-policy.github.io}}

\title{\LARGE \bf
The Ingredients for Robotic Diffusion Transformers
}

\author{
  Sudeep Dasari$^{1}$, Oier Mees$^{2}$, Sebastian Zhao$^{2}$, Mohan Kumar Srirama$^{1}$, Sergey Levine$^{2}$
  \thanks{
$^{1}$Carnegie Mellon University.%
$^{2}$University of California, Berkeley. %
}
}

\begin{document}

\maketitle
\thispagestyle{empty}
\pagestyle{empty}

\begin{abstract}
In recent years roboticists have achieved remarkable progress in solving increasingly general tasks on dexterous robotic hardware by leveraging high capacity Transformer network architectures and generative diffusion models. Unfortunately, combining these two orthogonal improvements has proven surprisingly difficult, since there is no clear and well-understood process for making important design choices.  
In this paper, we identify, study and improve key architectural design decisions for high-capacity diffusion transformer policies. The resulting models can efficiently solve diverse tasks on multiple robot embodiments, without the excruciating pain of per-setup hyper-parameter tuning. By combining the results of our investigation with our improved model components, we are able to
present a novel architecture, named \method, that significantly outperforms the state of the art in solving long-horizon ($1500+$ time-steps) dexterous tasks on a bi-manual ALOHA robot.
In addition, we find that our policies show improved scaling performance when trained on 10 hours of highly multi-modal, language annotated ALOHA demonstration data. We hope this work will open the door for future robot learning techniques that leverage the efficiency of generative diffusion modeling with the scalability of large scale transformer architectures.
Code, robot dataset, and videos are available at: \website
    
\end{abstract}
\input{sections/01-intro}

\input{sections/02-rw}

\input{sections/03-methods}
\input{sections/04-results}

\input{sections/05-discussion}

\section*{ACKNOWLEDGMENT}
This research was partly supported by NSF under IIS-2150826 and ONR under N00014-20-1-2383, and SD's PhD was supported by the NDSEG fellowship. Finally, we'd like to recognize thoughtful feedback from Katie Kang, Homer Walke, Dibya Ghosh, Oleg Rybkin, Pranav Atreya, and other members of the UC Berkeley RAIL lab that greatly improved this paper.

\bibliographystyle{IEEEtran}
\bibliography{ref}

\end{document}

%% file: sections/01-intro.tex
\section{Introduction}
\label{sec:intro}

\begin{figure}[t]
    \centering
    \includegraphics[width=\columnwidth]{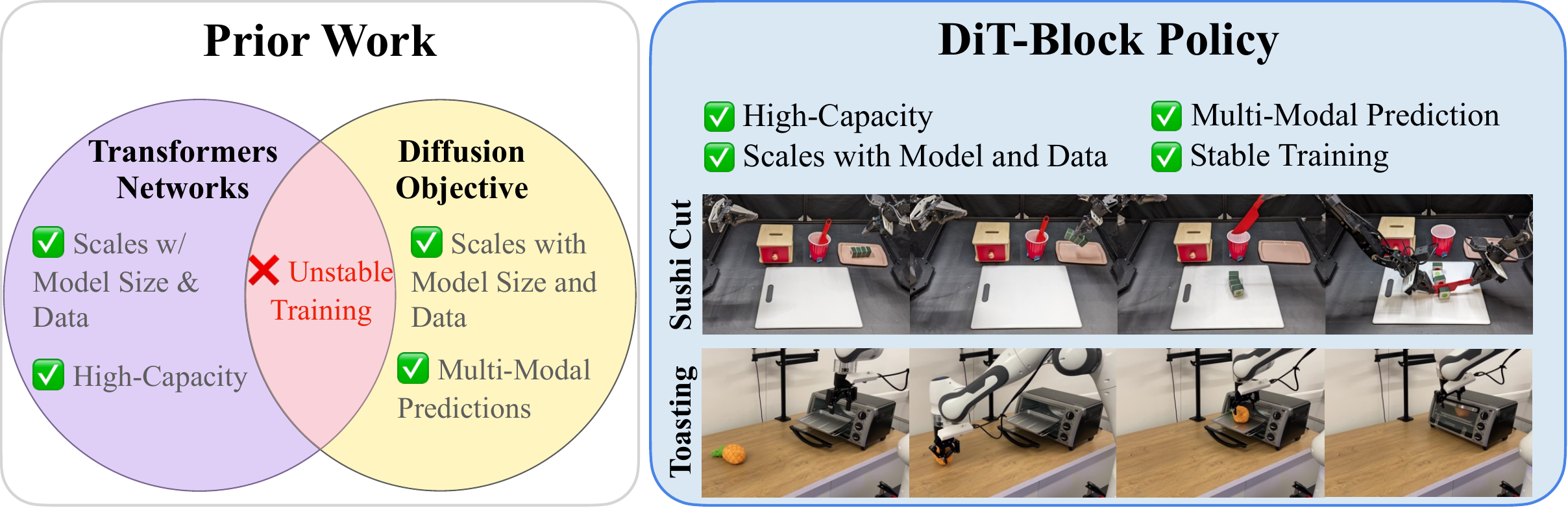}
    \caption{\textbf{Overview:} We introduce Diffusion Transformer Block Policies (i.e., \methods), a novel architecture that combines the scalability of Transformer backbones with generative modeling, without the excruciating pain of per-setup hyper-parameter tuning.}
    \label{fig:teaser}
    \vspace{-0.7cm}
\end{figure}

Modern machine learning has achieved remarkable success by leveraging highly expressive deep neural networks to generate and model samples from extensive offline imitation datasets~\cite{achiam2023gpt,esser2024scaling,kirillov2023segment}. Inspired by these advances, the field of robotics is adopting similar techniques to develop general policies and controllers for manipulation~\cite{octo_2023,open_x_embodiment_rt_x_2023} and locomotion tasks~\cite{shah2023vint,Doshi24-crossformer}. However, robotics tasks present multiple challenges that hinder the straightforward application of these methods. 
First, the policy must learn to process high-dimensional observation streams from multiple cameras, without overfitting to spurious correlations in the data. For example, the policy may learn to regress actions directly from proprioceptive signals or a specific camera view. Thus, during test time it would entirely ignore signals from other modalities (e.g., wrist cameras) that are critical for solving highly dexterous tasks with potential occlusions.
This often results in catastrophic failure during deployment. 
Second, the robot must make extremely precise action predictions, due to the low error tolerance in object manipulation. This is especially important when solving long horizon tasks, where the robot may need to achieve multiple sub-goals in sequence before the trajectory ends. For example, a robot tasked with preparing a sushi dish would need to reach multiple ``cutting'' sub-goals, which each have millimeter level error thresholds, as showcased in Fig.~\ref{fig:teaser}.
Finally, policy learning needs to contend with multi-modal action distributions -- \ie different ways of solving the same task. Simply learning the average action from this distribution will often result in an indecisive and error-prone behaviors. Handling action multi-modality becomes particularly important as the dataset size increases, since different experts will naturally demonstrate different behaviors. Failing to address these challenges will result in an unreliable and unsafe policy during deployment.

Recent advancements have begun to address these issues, by developing higher-capacity network architectures for dexterous task~\cite{zhao2023learning}, and  leveraging improved generative modeling frameworks like diffusion~\cite{sohl2015deep} for effective multi-modal action learning~\cite{chi2023diffusionpolicy}. 
Combining these two orthogonal improvements could yield highly capable policies, but has proven surprisingly challenging so far. For example, the original diffusion policy paper~\cite{chi2023diffusionpolicy} proposed a na\"ive cross-attention Transformer~\cite{vaswani2017attention} implementation for the policy network that was (according to their own analysis) extremely difficult to train. As a result, most follow-up works~\cite{prasad2024consistency} build upon their U-Net architecture~\cite{ronneberger2015u}, which is easier to tune but imposes strict requirements on the task setup (e.g., action signals must be sufficiently smooth~\cite{chi2023diffusionpolicy}). As a result, high-capacity diffusion modeling remains inaccessible for a wide range of robotics applications.

This work's key insight is that unstable transformer diffusion policy training is not a fundamental problem, and can be largely resolved with a novel policy architecture.
Our contributions are:
     \textbf{(1) Scalable Attention Blocks}: we propose a key improvement (inspired by Peebles et. al.~\cite{peebles2023scalable}) to stabilize training by adding adaptive Layer Norm (adaLN) blocks to the diffusion transformer policy layers. This simple trick improves performance by $30\%+$ on long horizon, dexterous, real-world manipulation tasks containing over 1000 decisions!
      \textbf{(2) Efficient Observation Tokenization}: we compare several methods to tokenize multiple camera observations, such as Vision Transformers~\cite{dosovitskiy2020image} and ResNet~\cite{he2016deep} encoders. Again, we find that a relatively simple implementation (ResNet image tokenizer + Transformer policy) can provide a substantial ($40\%+$) performance boost over competing strategies.
      \textbf{(3) \method:} We integrate the best performing components in a unified framework, coined \method. Our model achieves State Of The Art (SOTA) performance on both a bi-manual, low-cost ALOHA~\cite{zhao2023learning} robot, and on a single-arm DROID Franka setup~\cite{khazatsky2024droid}. 
       \textbf{(4) Open Source Models and Data:} We open-source all of our data, code and models for the community's benefit. This includes \datasetname, a new language annotated dataset containing 7023 clips of dexterous, bi-manual manipulation tasks %

%% file: sections/02-rw.tex
\section{ Related Work}
\label{sec:rw}

\paragraph{Encoding high dimensional observations} In order to perceive their environment, robots typically make use of multiple sensory observations. Therefore, how to best combine information from multiple sensors is a age-old question in robotics and computer vision~\cite{simonyan2014two,srivastava2012multimodal,hariharan2014simultaneous,mees16iros,eitel2015multimodal}. 
For example, bi-manual robots like ALOHA~\cite{zhao2023learning} must combine information from global cameras that view the whole scene and wrist cameras that get a close-up view of the manipulation itself. The most straight-forward way to handle this problem is to learn a single shared network/encoder that operates across all the input modalities simultaneously~\cite{octo_2023,robomimic2021}. However, these systems often learn brittle features that overfit to specific inputs, \eg proprioceptive data and global cameras, while ignoring others entirely. Possible solutions from prior work include using separate high-capacity image encoders for each visual stream~\cite{zhao2023learning,chi2023diffusionpolicy}, injecting 3D aware spatial biases into the representation network~\cite{ichnowski2021dex,kerr2022evo}, and properly regularizing the features using observation dropout~\cite{srivastava2014dropout,dasari2023datasets}.  Our findings reveal that a combination of these tricks provide a roughly $40\%$ boost on long-horizon, bi-manual tasks, and that these seemingly small details are crucial for successful visuo-motor control.

\paragraph{Predicting multi-modal action distributions} Modeling multi-modal action distributions -- \ie scenarios where the robot could take multiple entirely different actions from the same observation/goal -- is a well known challenge for BC methods~\cite{ross2011reduction}. This challenge often intensifies as the amount of expert data increases, since different demonstrations may showcase different solutions for the same task. Potential solutions include action space discretization~\cite{lynch2019play,dasari2021transformers,mees23hulc2,mees2022hulc,rosete2022corl}, modifying $\pi$ to predict higher capacity action distributions~\cite{robomimic2021,shafiullah2022behavior,rahmatizadeh2018virtual}, implicitly modeling the action distribution~\cite{florence2022implicit,song2021ebm,wang2024diffail,lai2024diffusion,chendiffusion}, and using a generative modeling objective like diffusion~\cite{ho2020denoising,song2019generative,chi2023diffusionpolicy,pearceimitating,saxena2023constrained,reuss2023goal}. Diffusion in particular has shown state-of-the-art  results in robotics~\cite{chi2023diffusionpolicy}: it can learn complex 3D-aware policies~\cite{ke20243d,ze20243d}, and concurrent work even showed state-of-the-art manipulation results on bi-manual robotic arms~\cite{zhao2024unleashed}. However, the model architectures/hyper-parameters are very sensitive and difficult to tune~\cite{chi2023diffusionpolicy,prasad2024consistency}. This is a major barrier to scaling, since higher-capacity network architectures, such as Transformers~\cite{vaswani2017attention}, are crucial to fitting large and more diverse datasets. In contrast, our approach alleviates these issues by replacing the standard cross/joint attention conditioning blocks in a transformer decoder, with one better suited for diffusion~\cite{peebles2023scalable}.

%% file: sections/03-methods.tex
\begin{figure*}[t]
    \centering
    \includegraphics[width=0.8\textwidth]{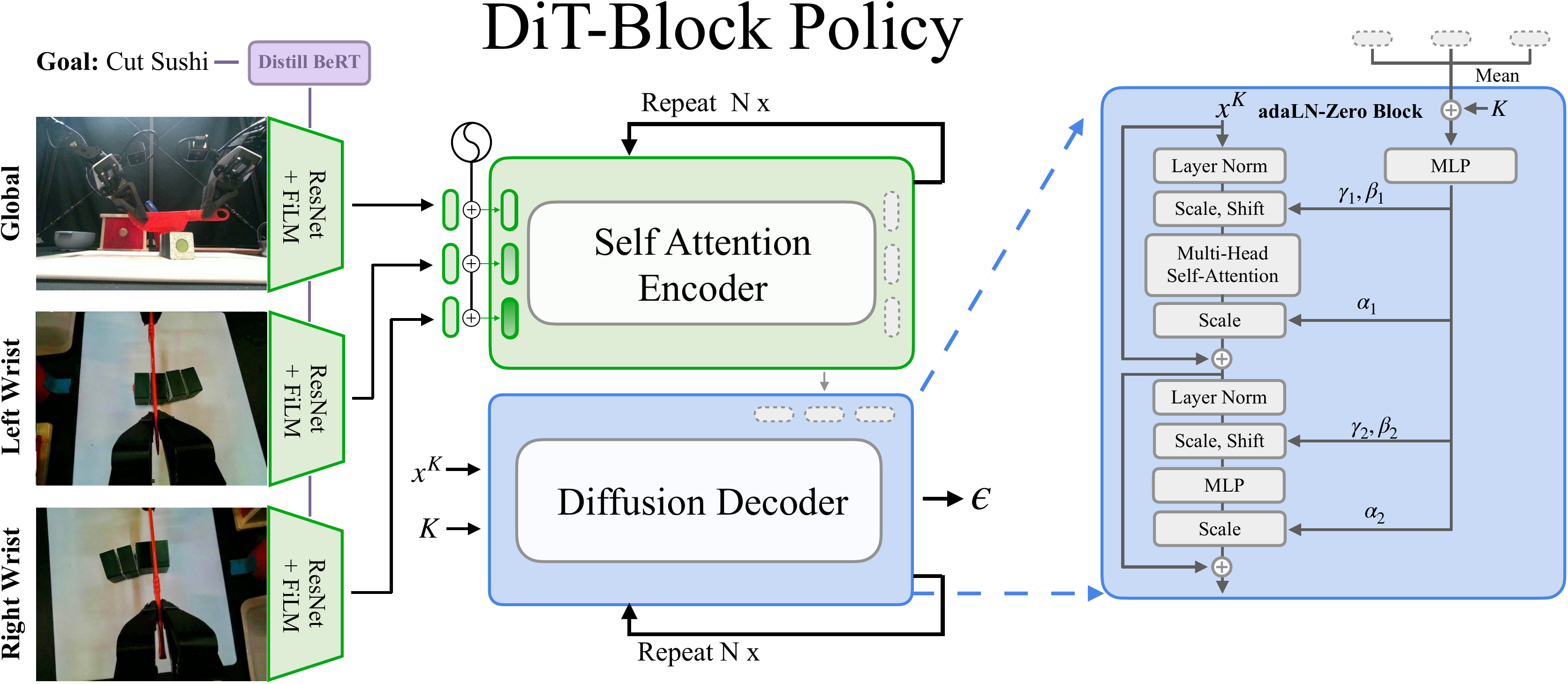}
    \caption{\textbf{Policy Architecture:} Our \method architecture enables scalable, goal-conditioned policy learning for various robotics tasks. Image observations are tokenized using separate ResNet-26~\cite{he2016deep} encoders. The text goal is tokenized and encoded into an embedding vector using a pre-trained Distill BeRT model~\cite{sanh2019distilbert}. This vector is incorporated into the observations tokens using FiLM Layers~\cite{perez2018film}. The observation tokens are passed into a encoder-decoder transformer network (middle), which is responsible for predicting the noise epsilon ($\epsilon$) used for diffusion. For stable training, the decoder block leverages a custom adaLN-Zero architecture (right), enabling the transformer to scalably optimize the diffusion objective.}
    \vspace{-0.6cm}
    \label{fig:arch}
\end{figure*}

\section{Problem Setting}
\label{sec:prob}

We consider the problem of acquiring a robotic controller via imitation learning~\cite{argall2009survey,billard2008survey,schaal1999imitation,kang2018policy,hester2018deep,weihs2021bridging} that can perform challenging, dexterous manipulation behaviors when prompted to via language instructions.
Specifically, the robot must learn a goal-conditioned policy $ \pi_{\theta} \left(a_t \mid o_t, g \right)$ that predicts an action distribution $a_t \sim \pi(\cdot | o_t, g) $, given a new input observation ($o_t$) and a desired language goal ($g$),
under environment dynamics $ \mathcal{T}: \mathcal{S} \times \mathcal{A} \rightarrow \mathcal{S}$, with $o_t \in \mathcal{S}$ and $a_t \in \mathcal{A}$. The policy $\pi$ is optimized via behavioral cloning~\cite{ross2010efficient,ross2011reduction,pomerleau1988alvinn} (BC) to match the optimal action distribution given a demonstration dataset $\mathcal{D} = \{ \tau_1, \dots, \tau_n \}$, where each trajectory $\tau_i = \{g, o_0, a_0, o_1, \dots\}$ was collected from an expert agent (e.g., human tele-op data). During test time, actions are sampled from $\pi$ and executed on the robot. We choose: $o_t$ to be a set of image observations from the robot; $a_t$ to be a chunk of $H$ joint/Cartesian state actions, and $g$ to be a text description of the task. This setting allows us maximum flexibility and generality for a wide range of robotics tasks, where precise states are difficult to infer and goals are free-form natural language instructions.

\subsection{Training Objective}
\label{sec:prob-training}
Our policy $\pi$ is formulated as a conditional Denoising Diffusion Probabilistic Model~\cite{ho2020denoising} (DDPM), a type of generative model where the output is sampled using a denoising process~\cite{welling2011bayesian}. Given the initial Gaussian noise $x^k$ and a noise prediction network $\epsilon_\theta(x^k, k, o_t, g)$ the DDPM process produces  $x^{k-1} = \alpha (x^k - \gamma \epsilon_\theta(x^k, k, o_t, g) + \mathcal{N}(0, \sigma^2 I))$, where $k$ is the diffusion time index and $\alpha, \gamma, \sigma$ are parameters associated with the diffusion noise schedule~\cite{ho2020denoising}. When $\epsilon_\theta$ is properly trained, this process will yield a sequence terminating in the optimal action: $x^k, x^{k-1}, \dots, x^0 \simeq a_t$. Thus, our goal is to learn $\epsilon_\theta$ via gradient descent~\cite{rumelhart1986learning,kingma2014adam} using the following MSE objective: $\mathcal{L} = || \epsilon^k - \epsilon_\theta(a_t + \epsilon^k, k, o_t, g)||_2^2$. Note that we use $k=100$ diffusion steps during training, a cosine noise schedule~\cite{nichol2021improved}, and a standard deterministic sampling process to reduce the number of samples needed (to $k=10$) during test time~\cite{song2020denoising}. %

\section{Introducing the \method}
\label{sec:components}

Our method -- \method -- is a Transformer neural network architecture designed specifically to be a highly performant conditional noise network ($\epsilon_\theta$ from above) for robotic diffusion policies. The \method architecture is visualized in Fig.~\ref{fig:arch}. First, the text goal and robot proprioception inputs are encoded into observation vectors. Similarly, the time-step $k$ is turned into an embedding vector using sinusoidal Fourier features~\cite{vaswani2017attention,tancik2020fourier} and a small MLP network. Then, all these embedding vectors are combined with the input noise vector ($x^k$) using an encoder-decoder Transformer architecture to produce the denoising output $\epsilon^k$. We now describe a few ingredients that are key to enable stable training and improved action prediction performance. %

\paragraph{Processing diverse multi-camera observations} Before passing through the transformer backbone, the input images, text goal, and joint angle observations need to be tokenized. The input images from each camera are processed \textit{separately}, using Convolutional Neural Network (CNN) backbones~\cite{xiao2021early}. While other vision transformers~\cite{dosovitskiy2020image,steiner2021train} may skip this stage entirely, the intensive spatial reasoning and limited data in many robotics tasks can benefit from the spatial priors in higher-capacity CNNs. Thus, we used ResNet-26~\cite{he2016deep} as the encoder.
The text goals are incorporated into the vision encoder via FiLM layers~\cite{perez2018film}. This enables the text goals to influence the network's visual attention at all layers of the network. Finally, the proprioceptive inputs are regularized with a per-dimension observation dropout~\cite{srivastava2014dropout,dasari2023datasets}, before tokenization. After the initial tokenization, learned positional encodings~\cite{vaswani2017attention} are added to the input tokens, and processed together using the Block Attention transformer encoder implementation from Octo~\cite{octo_2023}. These results in a series of transformer  joint embedding tokens $e^{(1)}, \dots, e^{(L)}$, where $L$ is number of layers.

\paragraph{Leveraging adaLN-Zero attention blocks for policy learning.} In parallel, a transformer decoder (with $L$ layers) processes the current (noised) input ($x^k$), time-step ($k$), and encoder embeddings. We note that each decoder block $i$ processes its corresponding embedding from the encoder $e^{(i)}$. Typically, this processing occurs via a standard cross attention mechanism, 
enabling the decoder to index into $e^{(i)}$ using its input tokens. Our key insight is, that this default attention implementation   
explains the poor training dynamics of prior diffusion policy transformer implementations~\cite{chi2023diffusionpolicy,prasad2024consistency}. 
Thus, we propose replacing standard cross-attention blocks with an adaptive Layer-Norm (adaLN) mechanism that plays a key role in stabilizing diffusion transformers in image generation tasks~\cite{peebles2023scalable}. These blocks work by injecting the conditioning vector into the Transformer's LayerNorm blocks, by shifting and scaling the input vectors: $x = a(e^{(i)}, k) * x + b(e^{(i)}, k)$. We choose $a$ and $b$ to be simple dense  layers that operate on the mean encoder embedding and the time vector: $a(e^(i), t) =  \texttt{tokenmean}(e^{(i)}) + t$.  
In addition, the output scales projection layers, before residual layer, are initialized to $0$ (hence adaLN-Zero).  This essentially initializes the noise network with identity skip connections, and thus further improves its learning dynamics~\cite{goyal2017accurate}.

%% file: sections/04-results.tex
\section{\methods for Bi-Manual Tasks}
\label{sec:datacol}
\begin{figure}[t]
            \vspace{0.2cm}
            \centering
            \includegraphics[width=0.85\columnwidth]{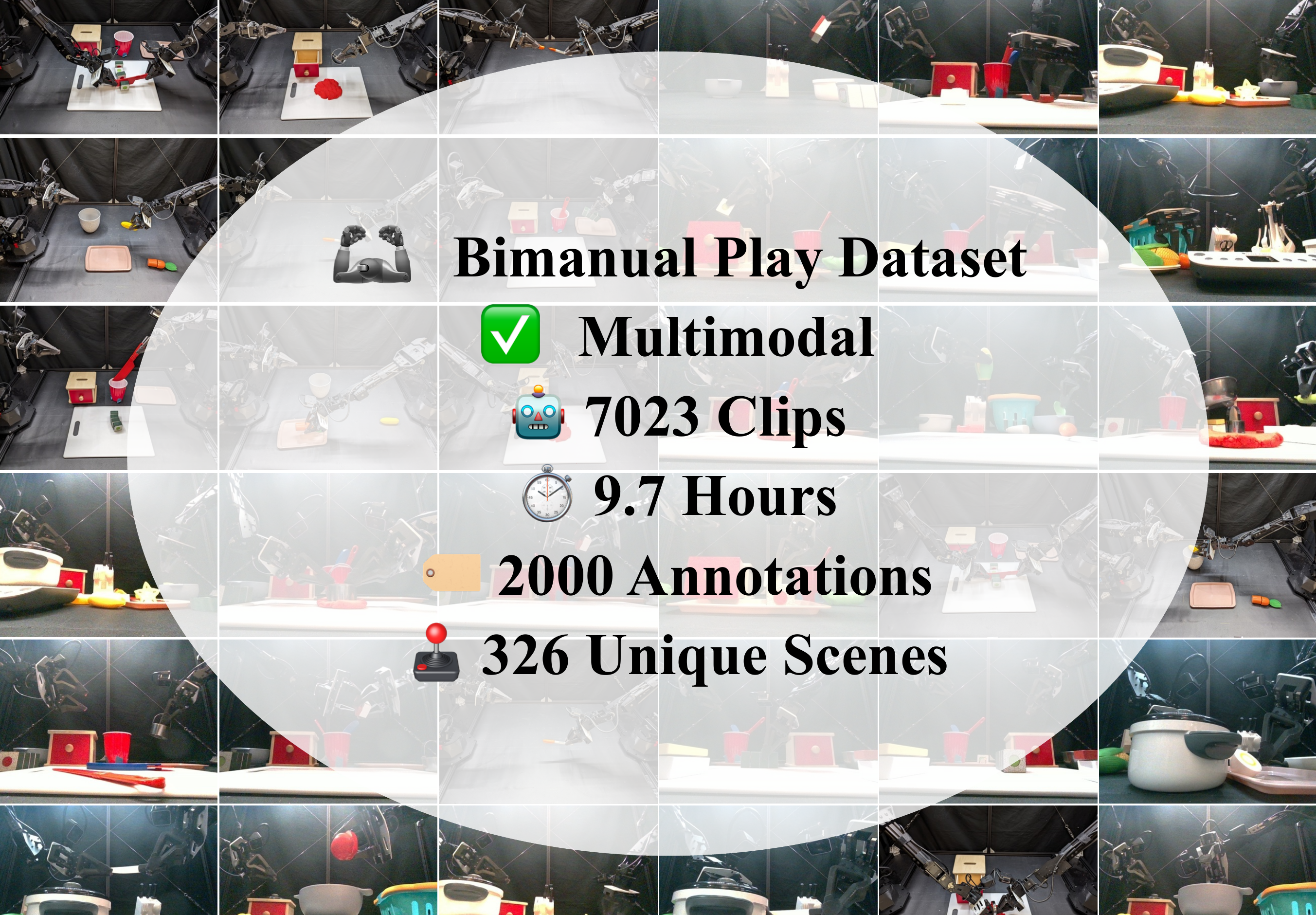}
            \captionof{figure}{
            \textbf{Introducing \datasetname:} To create this dataset we constructed 326 scenes in an ALOHA play-pen that we used to collect 7023 unique interaction sequences, with diverse objects, goals, language annotations and tasks. 
            }
            \label{fig:ourdata}
            \vspace{-0.6cm}
    \end{figure}

Inspired by prior work on data scaling~\cite{walke2023bridgedata,dasari2020robonet,khazatsky2024droid,open_x_embodiment_rt_x_2023,lynch2019play,mees2022calvin}, we seek to understand how \methods will behave as they are trained on increasingly diverse demonstrations data. However, %
the few (open-source) bi-manual datasets that do exist~\cite{zhao2023learning,shi2024yell} only consist of a handful of tasks, collected using the same controlled scenes/objects. As a result, they are not useful for testing generalization in our bi-manual setting. To address this shortcoming, we collected and annotated \datasetname, a more diverse bi-manual manipulation dataset with randomized objects and background settings as shown in Fig.~\ref{fig:ourdata}. We collected \datasetname as a series of 3.5 minute long episodes. For each episode, we constructed a random scene with various objects, and solved a sequence of tasks within that scene. After collection, the episodes were broken into clips that were in turn annotated with appropriate language task descriptions. The final dataset contains 7023 clips spanning 10 Hrs of robot data collection.%

\subsection{Training Protocol}
\label{sec:datacol-train}

To train our models, we collected a fixed set of demonstrations ($100+$ demos) for each of our evaluation tasks (see Sec.~\ref{sec:study-tasks}). In addition, we compiled open sourced data from prior work (ALOHA~\cite{zhao2023learning} dataset and optimal policy roll-outs from YaY~\cite{shi2024yell}), and added it to the training mix for regularization. The full mix of data is presented in Table~\ref{tab:trainmix}. All \methods were trained on this data-mix for 250K iterations, using the AdamW~\cite{kingma2014adam} optimizer and a cosine learning schedule~\cite{loshchilov2016sgdr}. Finally, instead of predicting a single action at each step, we trained \method models to predict a chunk of $H=100$ actions. This acted as regularization during training, and allowed us to employ temporal ensembling~\cite{zhao2023learning}, to improve stability at runtime.

\section{Experimental Setup}
\label{sec:study}

Our experiments are designed to understand \method's limits and capabilities. First, we defined a series of manipulation tasks using two different robot morphologies in Sec.~\ref{sec:study-tasks}. Then, we trained the policies on separate mixes of task demonstration data, grouped by morphology. %

\begin{figure}[b]
            \vspace{-0.2cm}
            \centering
            \setlength{\tabcolsep}{2.5pt}
                \begin{tabular}{lcccc}
                    \toprule
                     \textbf{Dataset} & \textbf{Make-Up} & \textbf{Scenes} & \textbf{Tasks} & \textbf{Length} \\
                     \midrule
                     \datasetname & 7k Play Clips & 326 & 200+ &  9.7 Hrs \\
                     \midrule
                     ALOHA~\cite{zhao2023learning} & 855 Demos & 15 & 16 &  2.9 Hrs \\
                     YaY~\cite{shi2024yell} & 4k Rollouts & 3 & 3 &   15.4 Hrs \\
                     \midrule
                     Pen Uncap & 100 Demos & 1 & 1 & 0.3 Hrs \\
                     Sushi Cut & 256 Demos & 1 & 1 & 2.7 Hrs \\
                     Pick Place & 863 Demos & 1 & 1 &  1.4 Hrs \\
                     Dough Cut & 150 Demos & 1 & 1 & 1.8 Hrs \\
                     Open Drawer & 115 Demos & 1 & 1 &  2.7 Hrs \\
                    \bottomrule
                \end{tabular}

            \captionof{table}{
            \textbf{Training Mix:} We train \method policies on: \datasetname, prior bi-manual manipulation datasets, and expert demonstration data collected for each task. %
            }
            \label{tab:trainmix}
    \end{figure}

\begin{figure*}[t]
    \vspace{0.2cm}
    \centering
    \includegraphics[width=0.9\textwidth]{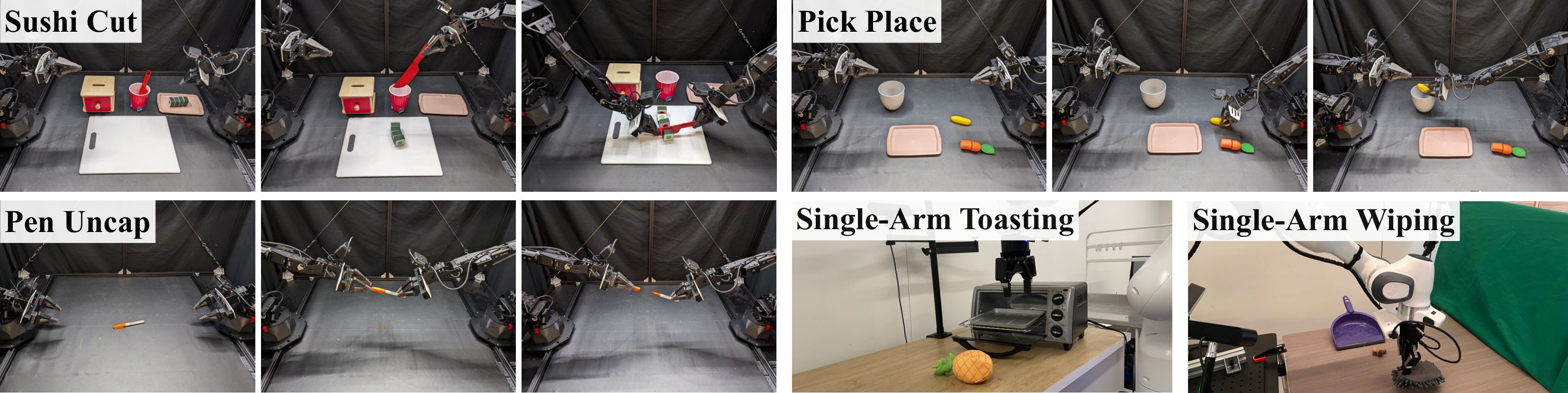}
    \caption{\textbf{Evaluation Tasks:} We evaluate \methods on a set of 3 Bi-Manual and 2 Single-Arm manipulation tasks. %
    }
    \label{fig:tasks}
    \vspace{-0.4cm}
\end{figure*}

\subsection{Task Setups}
\label{sec:study-tasks}

Our first task set considers a bi-manual, low-cost ALOHA robot~\cite{zhao2023learning}, which enables us to investigate challenging scenarios with highly dexterous, precise behaviors. We now describe these tasks and their success criteria in detail: \textbf{(1) Pick Place:} Given a text instruction (like $g=$``pick up the corn and place it in the bowl") the robot must find the target object, grasp it, and then drop it into the target plate/bowl. There are always two objects and two possible targets in the scene, so the robot must properly ground its behavior in the text instruction. A trial is marked successful if the object ends in the correct receptacle. \textbf{(2) Pen Uncap:}
This task evaluates the robot's precision grasping capabilities and its ability to control both arms simultaneously. The robot must pick up the sharpie with one arm, bring it to the other arm, and then remove the cap from the pen. A trial is marked successful if it ends with the pen uncapped.
\textbf{(3) Sushi Cut:} This task evaluated the robot's ability to chain precise, dexterous manipulation tasks over a long horizon. It is the most challenging task, since even a small error over a 2 min episode could derail the policy. The robot must place the sushi on the cutting board, pick up a knife from the cup, re-grasp it with the other hand, and then cut the sushi into four pieces. The task is marked successful if it ends with the sushi split into four, but partial credit is given for the fraction of successful cuts (e.g., one cut gets $1/3$).

Our next task set uses a single-arm Franka FR3 robot. While the dexterity is more limited, the Franka allows us to test generalization to an entirely new morphology and control space (Cartesian velocity). We consider the following tasks: \textbf{(1) Toasting:} In this long-horizon task, the robot must pick up the target object, place it in the toaster, and then shut the toaster. A trial is marked as successful if the toaster if the full sequence is completed, and is marked as half successful if the object is only placed in the toaster. \textbf{(2) Wiping:} The robot must localize the sponge, grasp it, and then push the debris into the dustpan. The trial is marked successful if all debris is wiped at the end of the run.

\section{Results}
\label{sec:results}
\input{tables/tab-base}
This section evaluates \method on our task suite  in order to contextualize its performance and analyze the source of its improvements. First, we compare \method against the strongest baselines from the field in Sec.~\ref{sec:results-baseline}, and find an average improvement of $20\%$. Next, the ablation studies (see Sec.~\ref{sec:results-ablate}) reveal that the diffusion head implementation is critical for stable training, and observation tokenizer architecture provides a significant performance boost. Finally, we show that these findings generalize to different robot hardware in Sec.~\ref{sec:results-droid}, and provide a standardized sim evaluation (see Sec.~\ref{sec:results-sim}). %

\subsection{Comparison to Prior SOTA Architectures}
\label{sec:results-baseline}

Our first experiments compare \method against SOTA baselines from the field in order to contextualize its performance. These baselines include: \textbf{(1) Action Chunking Transformers~\cite{zhao2023learning} (ACT):} ACT is built with a standard encoder-decoder transformer architecture, concretely DeTR~\cite{carion2020end}). The encoder processes input observation tokens, which include camera observations (encoded via ResNet-18~\cite{he2016deep}), goal conditioning vectors, and  (optionally) a latent plan vector computed from ground truth actions during training (randomly sampled during inference). The network is optimized via BC, using a L1-regression loss on expert actions. We implemented this baseline using the recommended hyper-parameters, and omitted the latent plan vector based on advice from the authors. In many respects, this model is analogous to \method, but with a standard attention block and no diffusion loss. \textbf{(2) Diffusion Policy w/ U-Net~\cite{chi2023diffusionpolicy} (D.P. U-Net):}  This is the original Diffusion Policy implementation from Chi et. al.~\cite{chi2023diffusionpolicy}. The camera observations are first processed into representation maps (via separate ResNets~\cite{he2016deep}), and then the dimensionality is reduced into a vector using spatial softmax~\cite{levine2016end}. This observation vector is then fed into a conditional U-Net network~\cite{ronneberger2015u} that functions as the noise network. The policy is trained using the DDPM diffusion training objective. \textbf{(3) Diffusion Policy w/ Transformer~\cite{chi2023diffusionpolicy} (D.P. Transformer):} This is the same setup as described previously, but with the U-Net noise network replaced with a Transformer, which uses a standard causal cross attention block. While higher capacity, this model is notoriously hard to train~\cite{chi2023diffusionpolicy,prasad2024consistency}.

All three baselines were compared against \method on our bi-manual evaluation tasks. Each method was trained twice, once with just the demonstration data and once with \datasetname, in order to understand their data scaling properties. Full results are presented in Table~\ref{tab:aloha-baselines}. We find that \method is able to outperform the strongest by roughly $20\%$ when trained with \datasetname, and by $10\%$ when trained on task data alone. This indicates that \method delivers SOTA performance, while also scaling better than the baselines. In addition, our method is able to deliver solid performance on all three tasks. In contrast, each of the other baselines has a task where it falls flat -- e.g., ACT struggles with pen uncap, and D.P. U-Net struggles with sushi cutting. Finally, note that the D.P. Transformer baseline is unable to solve \textit{any} of our tasks, because unstable training caused noisy/unsafe action prediction. Thus, we conclude that \method learns diffusion policy transformers more stably than the baseline does.
\subsection{Ablation Studies}
\label{sec:results-ablate}

\input{tables/tab-ablate-attn}
\paragraph{Ablating the attention mechanism} A key finding from the prior section is that \method's transformer implementation enables more stable training and policy inference. But is this inherent to the transformer architecture, or a factor of some other hyper-parameter? Thus, we conduct an apples-to-apples comparison in order to answer this question. We compare \method against 3 ablations that use the same exact setup, but with a different attention block: \textbf{(1) Cross Attention:} The diffusion decoder uses a standard per-layer cross attention block~\cite{vaswani2017attention} to condition on memory embeddings from the encoder stack (\ie ACT~\cite{zhao2023learning} + diffusion). Concurrent work~\cite{zhao2024unleashed} demonstrated SOTA results with this architecture, though with a much larger dataset (26K episodes) and extensive tuning. \textbf{(2) In-Context:} The memory embeddings from the encoder are added to the decoder in context, and all further processing happens with standard causal self-attention. \textbf{(3) Non-Zero Initialization:} This is an adaLN block, but without zero-initializing the final layers.

\input{tables/tab-ablate-feat}

We compare these ablations against a \method on the pick place and uncapping tasks. Results are presented in Table~\ref{tab:ablate-dit}. We find that the cross attention and in-context attention blocks are far less stable during training. It is still possible to generate stable actions during evaluation, by significantly increasing the number of diffusion steps during inference. However, the performance is still significantly reduced v.s. our \method, and the slow inference speed results in jerky trajectories when deployed on the robot. In contrast, we find that the zero initialization ablation is able to effectively train and predict actions with fewer inference steps. But it still under-performs the \method by $16\%$. Altogether, we conclude that the \method's architecture offers a critical boost for diffusion transformer policy performance, and that the initialization scheme provides an additional boost on top. 
\begin{figure}[b]
    \vspace{-0.5cm}
     \centering
     \includegraphics[width=0.8\columnwidth]{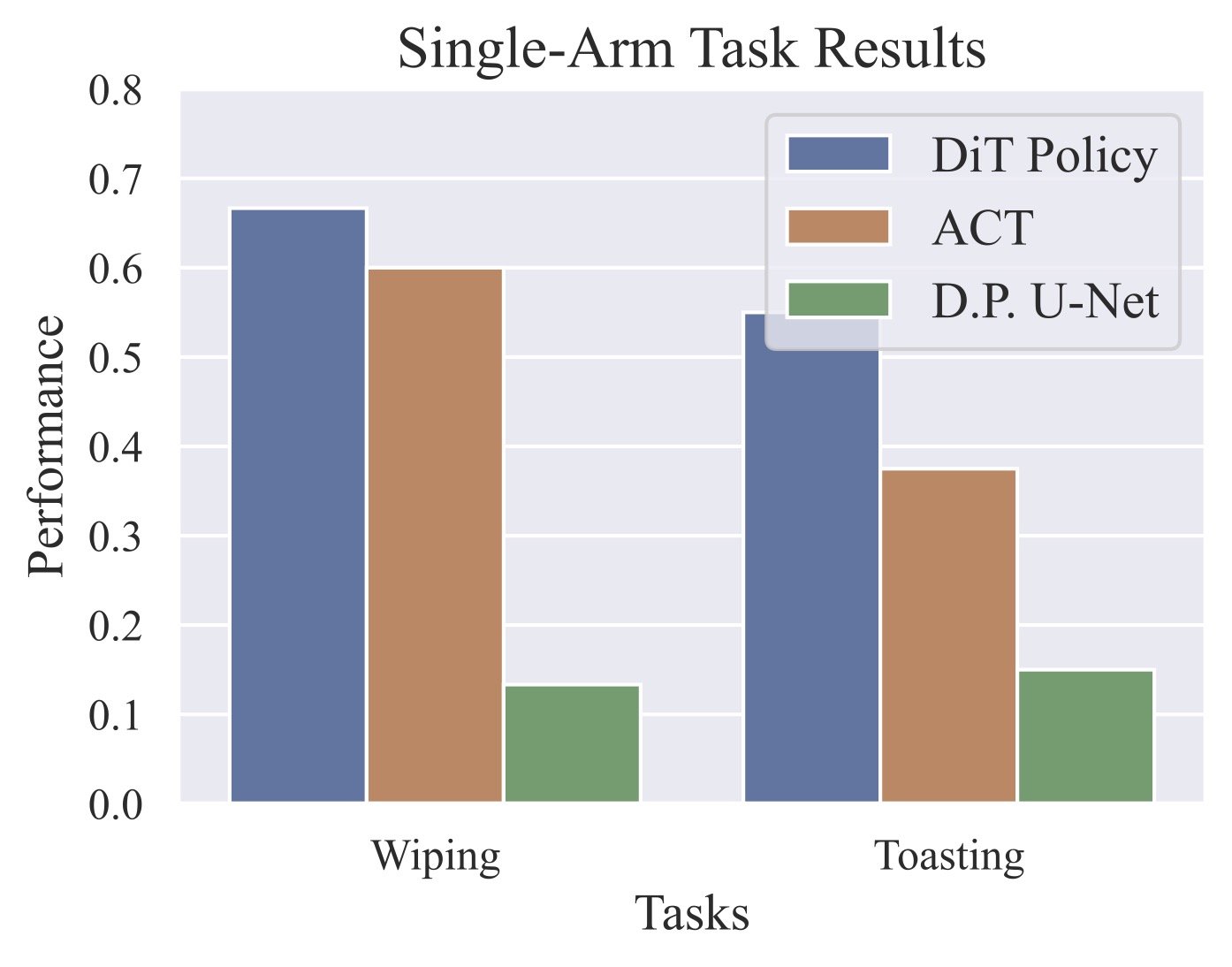}
     \caption{\textbf{Single Arm Real World Tasks:} We evaluate our \methods on the Franka real world single arm tasks, and find that it outperforms the strongest baseline by over $20\%$. %
    }
     \label{fig:results-droid}
\end{figure}

\paragraph{Ablating the image tokenization scheme} We evaluate our method's observation tokenizer by testing against ablations that move these parameters into the transformer encoder itself. Specifically, the ResNets are replaced with three small convolutional stem layers~\cite{octo_2023,xiao2021early} that produce an equivalent number of tokens (49 per image). 
Then, we train using these ablated observation tokens, and scale up the parameters to compensate. 
Results comparing these ablations against the full \method are presented in Table~\ref{tab:ablate-img}. 
We find that \method significantly outperforms the ablation with an even parameter count, and that even the significantly scaling up ablation is unable to compensate. This suggests that CNNs should still be considered as encoders for robotics tasks, particularly for low-data regimes.
\begin{figure}[t]
    \vspace{0.2cm}
    \centering
    \setlength{\tabcolsep}{2.5pt}
        \begin{tabular}{lcc}
            \toprule
             \textbf{Task} & \textbf{\method} & \textbf{D.P. Transformer~\cite{chi2023diffusionpolicy}} \\
             \midrule
             Lift & $100\%$ & $100\%$\\
             Can & $98\%$ & $100\%$\\
             Square & $84\%$ & $100\%$\\
             Tool Hang & $72\%$ & $76\%$\\
             \midrule
             Avg (Sim) & $88.5\%$ & $ 94\%$\\
             Avg (ALOHA) & $60\%$ & $ 0\%$\\
            \bottomrule
        \end{tabular} 
    
    \captionof{table}{
    \textbf{Sim Evaluation:} We compare \method against the original D.P. Transformer implementation~\cite{chi2023diffusionpolicy} on 4 tasks from the robomimic simulation eval suite~\cite{robomimic2021}. %
    }
    \label{tab:app-sim}
   \vspace{-0.5cm}
\end{figure}

\subsection{Generalization to Other Robot Morphologies}
\label{sec:results-droid}
The final experiments test if our findings still generalize to a new robot embodiment. Specifically, we test generalization to a single-arm Franka robot. While this setup is morphologically simpler, there are a few important differences that could prove challenging in practice. First, we evaluate the policies with a single external camera so they will need to gracefully handle occlusion during manipulation. Second, these robots use a velocity action space, which may prove more difficult to learn. We evaluate the two strongest baselines against \method on the toasting and wiping tasks. Results are presented in Fig.~\ref{fig:results-droid}. Note that \method again provides SOTA performance on these tasks: it outperforms ACT by $20\%$ on average and D.P. U-Net by $35\%$. This suggests that \method can generalize to new robots and is not overly sensitive to the particular choice of action and observation space, unlike the Diffusion Policy U-Net~\cite{chi2023diffusionpolicy}. 
\vspace{-0.1cm}
\subsection{Standardized Evaluation in Simulation}
\label{sec:results-sim}
While real-hardware evaluations are the ultimate test, it is often still useful to compare methods on reproducible, simulated task settings. Thus, we evaluate our \method against the reference Diffusion Policy Transformer (D.P. Transformer) baseline from Chi et. al.~\cite{chi2023diffusionpolicy} on the robomimic simulated task suite~\cite{robomimic2021}. The results are reported in Table.~\ref{tab:app-sim}. We find that \method almost completely matches D.P. Transformer on the simulated tasks, despite doing almost no task specific tuning (unlike D.P. Transformer). In addition, our method heavily out-performs the baseline on the real world experiments, which should carry more weight given the sim-to-real evaluation gap~\cite{dasari2023datasets}. %

%% file: tables/tab-base.tex
\begin{table*}[t]
    \vspace{0.2cm}
    \centering

        \begin{tabular}{lcc|cc|cc|cc}
            \toprule
             & \multicolumn{2}{c}{\textbf{Pick Place}} & \multicolumn{2}{c}{\textbf{Pen Uncap}} & \multicolumn{2}{c}{\textbf{Sushi Cut}} & \multicolumn{2}{c}{\textbf{Average}} \\
             \textbf{Using \datasetname?} & Yes & No &  Yes & No  & Yes & No & Yes & No \\
             \midrule
             \textit{\method} & $\textbf{50\%}$ & $37.5\%$ & $\textbf{100\%}$ & $90\%$ & $\textbf{29\%}$ & $13\%$ & $\textbf{60\%} \pm 9\%$ & $51\% \pm 9\%$\\
             \textit{ACT}~\cite{zhao2023learning} & $37.5\%$ & $25\%$ & $40\%$ & $70\%$ & $21\%$ & $17\%$ & $33\% \pm 8\%$ & $37\% \pm 8\%$ \\
             \textit{D.P. U-Net}~\cite{chi2023diffusionpolicy} & $31.3\%$ & $18.8\%$ & $90\%$ & $70\%$ & $4\%$ & $0\%$ & $42\% \pm 9\%$ & $30\% \pm 9\%$ \\
             \textit{D.P. Transformer}~\cite{chi2023diffusionpolicy} & $0\%$ & $0\%$ & $0\%$ & $0\%$ & $0\%$ & $0\%$ & $0\% \pm 0\%$ & $0\% \pm 0\%$ \\
            \bottomrule
        \end{tabular} 
    
    \caption{\textbf{Baseline Comparison:} We compare \method against SOTA baselines from the field (ACT~\cite{zhao2023learning}, Diffusion Policy w/ U-Net~\cite{chi2023diffusionpolicy}, and Diffusion Policy with Transformer~\cite{chi2023diffusionpolicy}).  Our method is able to outperform the baselines by  $20\%$. }
    \vspace{-0.6cm}
    \label{tab:aloha-baselines}
\end{table*}

%% file: tables/tab-ablate-attn.tex
\begin{figure}[b]
    \vspace{-0.4cm}
    \centering
    \setlength{\tabcolsep}{2.7pt}
        \begin{tabular}{lccc}
            \toprule
             \textbf{Method} & \textbf{DDIM Iters.} & \textbf{Pick Place} & \textbf{Pen Uncap} \\
             \midrule
             Ours & $10$ & $\textbf{50\%} \pm 12\%$ & $\textbf{100\%} \pm 0\%$ \\
             \midrule 
             No Zero-Init. & $10$ & $38\% \pm 11\%$ & $80\% \pm 13\%$\\
             \midrule
             Cross Attn.~\cite{zhao2024unleashed} & $10$ & $0\% \pm 0\%$ & $0\% \pm 0\%$\\
             Cross Attn.~\cite{zhao2024unleashed} & $100$ & $38\% \pm 15\%$ & $70\% \pm 11\%$ \\
             \midrule 
             In Context & $100$ & $0\% \pm 0\%$ & $0\% \pm 0\%$ \\
            \bottomrule
        \end{tabular} 
    \captionof{table}{
    \textbf{Attention Block Ablation:} Our proposed attention architecture significantly improves v.s. baselines.
    }
    \label{tab:ablate-dit}
    \vspace{-0.1cm}
\end{figure}

%% file: tables/tab-ablate-feat.tex
\begin{figure}[h]
    \vspace{0.2cm}
    \centering
        \begin{tabular}{lccc}
            \toprule
             \textbf{Method} & \textbf{Parameters} & \textbf{Pick Place} & \textbf{Pen Uncap} \\
             \midrule
             Ours & $115M$ & $\textbf{50\%} \pm 12\%$ & $\textbf{100\%} \pm 0\%$ \\
             \midrule 
             No ResNet & $85M$ & $0\% \pm 0\%$ & $0\% \pm 0\%$\\
             No ResNet & $150M$ & $13\% \pm 8\%$ & $20\% \pm 13\%$\\
            \bottomrule
        \end{tabular} 
        
    \captionof{table}{
    \textbf{Encoder Ablation:} We ablate our choice of ResNet encoder tokenization, which effectively shifts more compute/parameters below the transformer layers. %
    \vspace{-0.5cm}
    }
    \label{tab:ablate-img}
    
\end{figure}

%% file: sections/05-discussion.tex
\section{Conclusion}

\label{sec:discussion}

This paper presents \method, an improved transformer architecture that enables stable diffusion policy learning and efficient inference. Our experiments show that \methods provide SOTA performance across 5 tasks and 2 different robots, which have radically different observation spaces, action spaces, and morpohologies. We find that \method outperform the strongest baselines by $20\%$, and are able to scale better with diverse play data. Our ablation study reveals that the exact configuration of \method's transformer block is responsible for this increase. Standard joint-attention mechanisms are simply not able to learn policies as stably as \method can. In addition, an ablation of our observation tokenizer reveals that using separate ResNet CNNs for image encoding provides stronger performance than using transformers alone. Even scaling the transformers is not enough to make up for this difference. Finally, we open-source the \datasetname dataset used in our experiments. This is the first language annotated, bi-manual dataset with diverse scenes, tasks, and objects.